\pdfoutput=1

\documentclass[11pt]{article}
\usepackage{titlesec}

\usepackage{multicol}  

\usepackage[final]{acl}
\usepackage{amsmath}
\usepackage{amssymb}

\usepackage{times}
\usepackage{latexsym}

\usepackage[T1]{fontenc}

\usepackage{graphicx}
\usepackage{multirow}
\usepackage{booktabs}
\usepackage{adjustbox}
\usepackage{caption}

\usepackage{xcolor,colortbl}

\usepackage{enumitem}

\usepackage{tabularx} 

\usepackage{array}
\usepackage{color}
\definecolor{codeblue}{rgb}{0.0, 0.0, 1.0} 
\definecolor{codeblack}{rgb}{0.0, 0.0, 0.0} 
\usepackage{ragged2e}

\usepackage[utf8]{inputenc}

\usepackage{microtype}

\usepackage{inconsolata}

\usepackage{CJKutf8}
\usepackage{tcolorbox}
\tcbuselibrary{breakable, xparse, skins}
\usepackage{xltabular} 
\usepackage{listings} 
\usepackage{float}

\setlist[itemize]{nosep, topsep=0pt, partopsep=0pt}

\usepackage[symbol]{footmisc}
\title{IPL: Leveraging Multimodal Large Language Models for Intelligent Product Listing}

\author{
\begin{minipage}[t]{\textwidth}
\centering
\textbf{Kang Chen\textsuperscript{*,1,2,†}} \quad
\textbf{Qingheng Zhang\textsuperscript{*,1}} \quad
\textbf{Chengbao Lian\textsuperscript{*,1}} \quad
\textbf{Yixin Ji\textsuperscript{1}} \quad
\textbf{Xuwei Liu\textsuperscript{1}} \\
\textbf{Shuguang Han\textsuperscript{1,‡}} \quad
\textbf{Guoqiang Wu\textsuperscript{1}} \quad
\textbf{Fei Huang\textsuperscript{1}} \quad
\textbf{Jufeng Chen\textsuperscript{1}} \\
\textsuperscript{1}Alibaba Group \quad
\textsuperscript{2}Fudan University \\
\end{minipage}
\\[2ex]
\begin{minipage}[t]{\textwidth}
\centering
\texttt{
kchen24@m.fudan.edu.cn, \{qingheng.zqh, lianchengbao.lcb, jiyixin.jyx, xuweiliu.lxw, shuguang.sh, kingwu.wgq, jufeng.cjf\}@alibaba-inc.com
}
\end{minipage}
}

\newcommand{\newshortname}[1]{#1}

\begin{document}

\maketitle

\begin{abstract}

\renewcommand{\thefootnote}{\fnsymbol{footnote}}
\footnotetext[1]{These authors contributed equally to this work.}
\footnotetext[2]{Work done during an internship at Alibaba Group.}
\footnotetext[3]{Corresponding author: Shuguang Han (email: shuguang.sh@alibaba-inc.com)}

\renewcommand{\thefootnote}{\arabic{footnote}}

\renewcommand{\thefootnote}{\arabic{footnote}}

Unlike professional Business-to-Consumer (B2C) e-commerce platforms (e.g., Amazon), Consumer-to-Consumer (C2C) platforms (e.g., Facebook marketplace) are mainly targeting individual sellers who usually lack sufficient experience in e-commerce. Individual sellers often struggle to compose proper descriptions for selling products. With the recent advancement of Multimodal Large Language Models (MLLMs), we attempt to integrate such state-of-the-art generative AI technologies into the product listing process. To this end, we develop IPL, an Intelligent Product Listing tool tailored to generate descriptions using various product attributes such as category, brand, color, condition, etc. IPL enables users to compose product descriptions by merely uploading photos of the selling product. More importantly, it can imitate the content style of our C2C platform Xianyu\footnote{Xianyu is the largest C2C e-commerce platform in China.}. This is achieved by employing domain-specific instruction tuning on MLLMs and adopting the multi-modal Retrieval-Augmented Generation (RAG) process. A comprehensive empirical evaluation demonstrates that the underlying model of IPL significantly outperforms the base model in domain-specific tasks while producing less hallucination. IPL has been successfully deployed in our production system, where 72\% of users have their published product listings based on the generated content, and those product listings are shown to have a quality score 5.6\% higher than those without AI assistance.
\end{abstract}

\section{Introduction}

\begin{figure}[ht]
  \centering
  \captionsetup{skip=5pt}
  \includegraphics[width=0.46\textwidth]{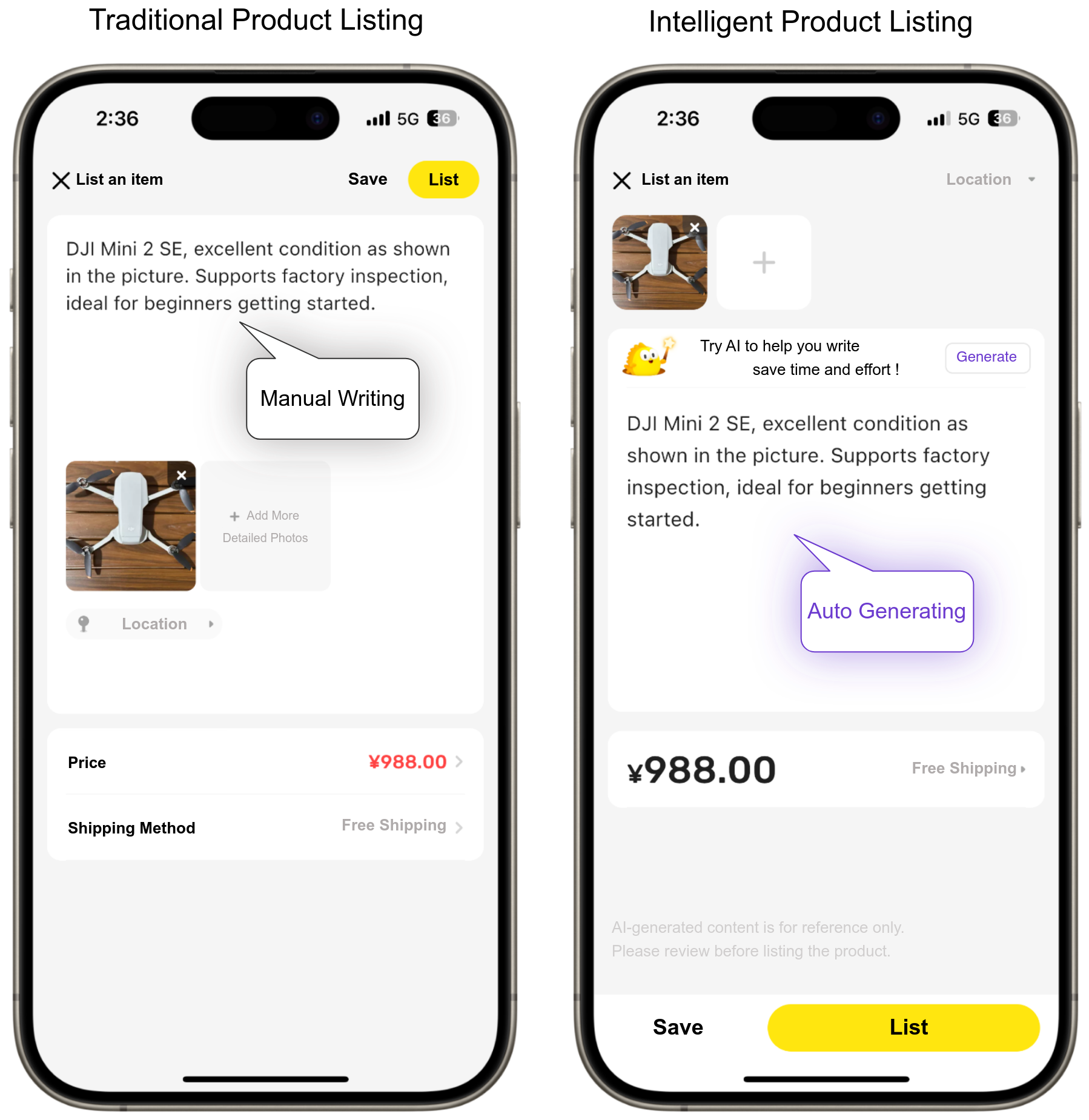} 
  \caption{Intelligent Product Listing on C2C Platforms}
  \label{fig:intro}
  \vspace{-10pt}
\end{figure}

With the rise of the circular economy, second-hand e-commerce has played a vital role in our daily lives. Unlike Business-to-Consumer (B2C) e-commerce (e.g., Amazon, Walmart), second-hand e-commerce is often operating in the form of Consumer-to-Consumer (C2C) transactions. Different from professional sellers on B2C platforms, individual sellers in second-hand marketplaces are usually inexperienced. They face unique challenges when listing their products --- navigating through the complicated listing procedure, and creating high-quality product descriptions. These issues not only affect the success rate of product listings but also impact the overall quality and discoverability of the listed products.

To address the above issues, it is imperative to simplify the listing process for individual users by leveraging automation to generate high-quality product descriptions. A typical product listing process involves users manually filling in basic product attributes, uploading product photos, and composing content descriptions. Among these steps, preparing product photos is relatively straightforward. If we can automatically generate product descriptions based on the uploaded photos, it would significantly reduce the listing effort and enhance user experience, as illustrated by Figure~\ref{fig:intro}.

Fortunately, product photos contain a wealth of information, enabling us to infer basic attribute information such as category, brand, and model from the imagery in most cases. Moreover, recent advancements in Multimodal Large Language Models (MLLMs) \cite{bai2023qwen2,achiam2023gpt} have significantly improved both visual understanding and natural language generation capabilities, making it feasible to generate product descriptions based on product photos in an automatic manner.

Several large e-commerce platforms, including eBay \cite{herold2024lilium} and Amazon \cite{jiang2024hallucination}, have begun to explore this direction by introducing product listing assistants. However, these tools are still in their infant stages. They still require substantial user input, and the generated content is commonly in the professional marketing styles which lowers the information authenticity for a C2C platform. In the context of second-hand e-commerce, we encounter more challenges.

\textbf{Lack of Domain Knowledge.} To generate high-quality product descriptions, models must possess strong capabilities for domain understanding~\cite{escursell2021sustainability,poerner2019bert}. C2C e-commerce differs from traditional B2C platforms, its product listings often exhibit more unique and varied characteristics. Unlike professional marketing descriptions that emphasize persuasive language, product descriptions in C2C platforms typically exhibit a more colloquial style, focusing on information authenticity. This helps foster trust between buyers and sellers and potentially facilitates transactions. However, existing MLLMs often fall short in these areas.

\textbf{Hallucination Problem.} Ideally, users only need to upload a photo, and the corresponding content description including core product attributes is automatically generated. However, achieving this goal imposes a significant challenge on the current MLLMs~\cite{liang2022holistic,ji2023survey}. In practice, MLLMs sometimes produce product attributes going beyond the image itself. This is known as the hallucination problem in Large Language Models(LLMs). As the core part of the product listing experience, we need to find a proper solution.

\textbf{Challenges for Production Deployment.} Deploying generative LLMs on production systems, particularly for applications with a large-scale user base, imposes high requirements on system latency, cost consumption~\cite{kwon2023efficient}, and content safety~\cite{perez2022ignore}. Meeting these demands necessitates a comprehensive system engineering effort.

To address the above issues, we develop an \textbf{I}ntelligent \textbf{P}roduct \textbf{L}isting (\textbf{IPL}) system, aiming to improve the efficiency and effectiveness for product listings on our production system. 

Firstly, we present a notable case study of injecting domain knowledge into an MLLM through further instruction tuning of an open-source model. Our domain-specific model significantly enhances the base model's understanding of domain knowledge and enables it to generate product descriptions in the unique style characteristic of C2C platforms. 

Secondly, we introduce an innovative multi-modal Retrieval-Augmented Generation (RAG) approach for visual-based content generation, leveraging identical product retrieval, to enhance description quality and mitigate hallucination risks in practical applications.

Finally, we have successfully deployed the system in our production system, delivering intelligent composition service to real-world individual users. This system demonstrates high user acceptance and effectively enhances the efficiency and quality of product listings.

Our extensive empirical studies demonstrate that IPL has the potential to transform the landscape of product listings, offering a robust, scalable solution to challenges faced by individual sellers and platforms alike.

\begin{figure*}[ht]
  \centering
  \includegraphics[width=0.8\textwidth]{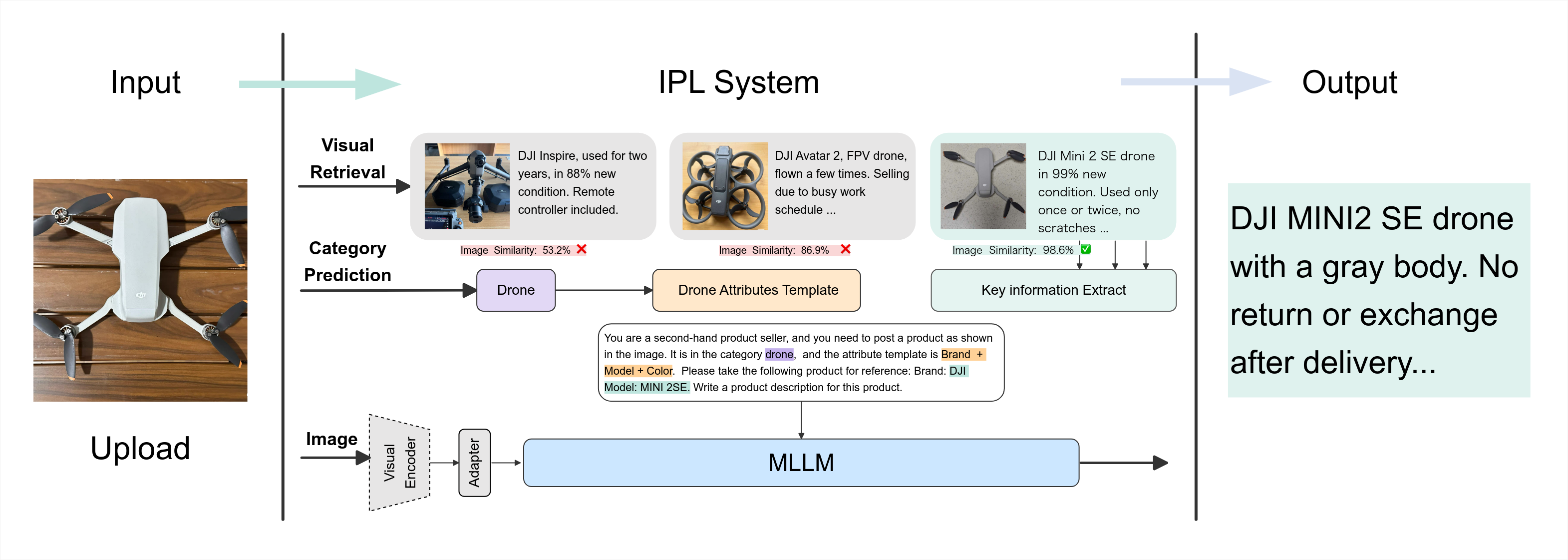} 
  \caption{Overview of the Intelligent Product Listing (IPL) system architecture.}
  \label{fig:example}
  \vspace{-10pt}
\end{figure*}

\section{Approach}
The overall architecture of our intelligent product listing system can be illustrated in Figure 2, which comprises an online multi-modal Retrieval-Augmented Generation (RAG) process for identifying similar products, and an offline-trained domain-specific MLLM for product description generation. 

In our product listing system, user-uploaded photos will go through several sub-modules: category prediction, retrieval of similar products, and extraction of key attributes (e.g., brand, model, etc.) from the descriptions of these similar products. Subsequently, the product photo, category, and extracted attributes are fed into the domain-specific MLLM as contextual information to generate the product description. With this automatically generated description, users only need to make minimal adjustments to complete the product listing.

\subsection{Domain-Specific Model Training} 
The crucial stages in training domain-specific models include the construction of training data and the process of model instruction tuning.

\subsubsection{Domain Instruction Tuning Data}

The training data for the model encompasses product description generation, domain content understanding, and general instruction tasks. The general instruction data are derived from both automatically generated and open-source data. An overview of the training data is provided in Table 1.

\begin{table}[H]
\centering
\resizebox{\columnwidth}{!}{
\begin{tabular}{@{}lccccccccc@{}}
\toprule

\textbf{Data Type} & \textbf{Size} & \textbf{Source} & \textbf{Modality} \\
\midrule
Product Description Generation & 267k & In-house  & Visual-Language \\
Domain Content Understanding & 200k & In-house  & Visual-Language, Text Only \\
Auto Generated Datasets & 378k & In-house & Visual-Language \\
General QA Datasets & 424k & Open source  & Visual-Language, Text Only \\
\textbf{ALL} & \textbf{1.27M} & Mixture  & Visual-Language, Text Only \\
\bottomrule
\end{tabular}
}
\caption{Instruction tuning training data}

\end{table}

The description generation dataset, which constitutes the primary focus of this work, involves generating descriptions based on user-provided product photos. By cleaning data from actual user-posted product listings, we obtained pairs of product photos and descriptions. Subsequently, we converted the data into various types of instruction formats, including generating product descriptions directly from photos and generating descriptions based on a combination of product photos, key attribute templates, and reference information, as illustrated in Table 2. Detailed data construction procedures are provided in Appendix A.1.

The content understanding tasks primarily include fundamental tasks in e-commerce scenarios, especially those on C2C platforms, aimed at enhancing the model's domain knowledge. These tasks include product image category prediction, product attribute extraction, and text similarity matching, among others. This data is derived from manually annotated data accumulated over time in business scenarios. Further details on the data can be found in Appendix A.2.

Finally, the general instruction dataset are used to enable the model to retain general capabilities and enhance its generalization ability. We employ large language models to generate general instructions and answers in the native language based on product photos, while also incorporating high-quality open-source academic datasets as supplementary resources. For further details, please refer to Appendix A.3.

\subsubsection{Model Training}
We chose Alibaba's Qwen-VL\cite{bai2023qwen2} model as the base model, primarily due to its strong performance in the native language and its robust open-source ecosystem. We employed full-parameter fine-tuning for model training, freezing the visual encoder module while updating the VL-Adapter and LLM components only (7B parameters).

The training objectives focused on classic next token generation for language model optimization, specifically excluding loss calculation for prompt prefixes and focusing on the special markers and the model output tokens. The objective can be formally defined as:

\begin{equation}
    L = -\sum_{t=1}^{T} \log P(y_t | y_{<t}, X)
\end{equation}

where \( X \) denotes the model input instructions, \( y \) represents the generated tokens, \( t \) refers to the position within the generated sequence, and \( T \) is the length of the final generated sequence. Further training details can be found in Appendix B.

\subsection{Online Retriever-Augmented Generation}

In the online phase, the fine-tuned domain model is capable of generating descriptions for product photos. To further mitigate hallucinations, our instructions are not to directly generate descriptions from product photos but to refer to product categories, core attribute templates, and retrieved information, as detailed in Table 2.

\begin{table}[h]
\centering
{\footnotesize 
\begin{tabular}{|>{\color{codeblack}\ttfamily}m{7cm}|}
\hline
\textcolor{codeblue}{\textbf{Generation with Reference Information}} \\ \hline
\textcolor{codeblue}{Prompt:} You are an experienced seller on a second-hand trading platform and need to post a \textbf{cell phone category} with the product image as shown in the picture, and the copy template is \textbf{Brand + Model + Storage Capacity + Color + Version + Screen Condition}. In which, \textbf{the brand is Huawei, the model is Mate10pro, the storage capacity is 6+64GB}, please write a paragraph description for this product. \\ \hline
\textcolor{codeblue}{Response:} \\
Personal used Huawei Mate10pro 6+64GB, Blue, condition as shown in the pictures, Mainland China version, screen in perfect condition without aging or scratches, all original, for those interested, please contact me privately. \\ \hline
\end{tabular}
} 
\caption{Instruction for product description generation with Retriever-Augmented Generation.}
\vspace{-10pt}
\end{table}

Therefore, in online scenario, product description generation is a Retrieval-Augmented Generation (RAG) process. We conduct category prediction on the input product photos and simultaneously retrieve identical products through vector retrieval. From the retrieved products, we extract key attribute values to serve as reference information for generating descriptions. The extraction of key attribute values is accomplished using a domain-specific large model we trained, with the prompt shown in Table 3. Key attribute sets for each category are derived from offline mining and manual summarization, and can be retrieved through product category queries. By incorporating attributes template into the instructions, we can further control the attributes and their sequence that the model must mention in the generated product descriptions, ensuring the richness of the information in the output descriptions.

\begin{table}[!htbp]
\centering
{\footnotesize 
\begin{tabular}{|>{\color{codeblack}\ttfamily}m{7cm}|} 
\hline
\textcolor{codeblue}{\textbf{Attribute Extraction Example}} \\ \hline
\textcolor{codeblue}{Prompt:} Extract the Brand, Model, Storage Capacity, Color, Version, Screen Condition for the following smartphone product. Output the result in JSON format. Product description: Huawei mate10Pro 6+64G completely original unrefurbished smartphone Mainland China version light scratches. \\ \hline
\textcolor{codeblue}{Response:} \\
\begin{verbatim}
{
  "Brand": "Huawei",
  "Model": "mate10Pro",
  "Storage Capacity": "6+64G",
  "Version": "Mainland China"
}
\end{verbatim} 
\\ \hline
\end{tabular}
} 
\caption{Attribute extraction instruction examples.}
\vspace{-10pt}
\end{table}

The category prediction model utilizes the ALBEF network architecture\cite{li2021align,zhang2018visual}, a classic vision-language multimodal model. The model has been pre-trained on domain-specific data and fine-tuned with millions of manually annotated datasets, achieving an accuracy of over 80\% across tens of thousands of categories. The implementation of the visual search draws upon the work conducted by \cite{zhang2018visual}. We select the most similar result from the retrieval outcomes as the identical product and impose a similarity score threshold to further enhance the accuracy. In offline evaluations, the accuracy of image retrieval for identical products is over 60\%, and for similar products, it is over 90\%. For more details on the evaluation of visual retrieval, please refer to Appendix C.1.

\section{Deployment}

Key considerations for LLM deployment included minimizing online latency, ensuring user experience, and addressing safety risks associated with content generation. We deployed the system online, with the LLM model hosted on NVIDIA® Tesla® V100 machines. Through various acceleration techniques, such as model quantization, ViT operation optimization, key-value caching, kernel operation fusion, and parallel computation\cite{aminabadi2022deepspeed,dao2022flashattention,dao2023flashattention}, the overall pipeline's average response time (RT) was reduced from 5 seconds to below 3 seconds. The adoption of streaming output ensured user experience by reducing wait times.We perform preemptive risk assessment on user-uploaded product photos and security checks on generated descriptions to prevent non-compliant content, thereby effectively avoiding public opinion risks. For more detailed error detection and exception handling, please refer to Appendix D.

\section{Experiment}
\subsection{Data}
Our experimental data comprises both domain-specific and general datasets. All data were sourced from real e-commerce scenarios and the target labels were either manually annotated or confirmed by actual platform users, then converted into instruction format. We constructed validation datasets encompassing tasks such as sentiment analysis, information extraction, content topic selection, tagging/classification, and attribute-based visual question answering within the e-commerce domain (For more details, refer to Appendix E). Additionally, we included datasets specifically designed to evaluate generative style and hallucination.

\subsection{Model}
\textbf{Domain-Specific Models:} To assess the effectiveness of domain knowledge injection, we trained several models with varying amounts of training data. The datasets were randomly shuffled and truncated. The comparison models include: Qwen-VL (baseline, without domain training), 10\% Data (trained with 10\% of the data), 20\% Data, 50\% Data, and 100\% Data.\\
\textbf{Online RAG System:} In addressing hallucination alleviation, we conducted experiments on various components of our online RAG system. This included evaluating the use of product category information, reference information from identical or similar products.

\subsection{Metrics}
Our evaluation encompasses comprehensive metrics to assess different aspects of model performance:\\
\textbf{N-gram-Based Metrics}: We employed BLEU \cite{papineni2002bleu}, ROUGE and ROUGE-L \cite{lin2004rouge} to evaluate the alignment of generated text with ground truth product descriptions.\\
\textbf{Semantic Similarity Metrics}: BERT embeddings measured semantic similarity (SIM) between model outputs and ground truth using BERT-Score.\\
\textbf{Task-Specific Accuracy Metrics}: These metrics were used for domain-specific knowledge questions, assessing model accuracy in understanding and responding to task-specific prompts.\\
\textbf{Human Assessment:} Evaluation was conducted by experts in the C2C domain, assessing whether the generated descriptions adhere to domain-specific style and identify key attributes\cite{chen2024humans} accurately. We perform a quantitative analysis of the results.

\section{Results}

In the following subsections, we discuss the five key research questions regarding our domain model and the online RAG system:

\begin{itemize}
    \item \textbf{Q1:} Does the domain-specific model, after instruction tuning, exhibit a stronger understanding of domain knowledge?
    \item \textbf{Q2:} Does the domain-specific model generate product descriptions with a more distinct C2C domain style?
    \item \textbf{Q3:} Can the model maintain its general capabilities after being trained on domain-specific data?
    \item \textbf{Q4:} Does the online RAG mitigate hallucinations in product description generation?
    \item \textbf{Q5:} How does the IPL system perform in real-world online scenarios?
    
\end{itemize}
Among them, Q1-Q3 investigate the effects of domain knowledge injection, Q4 explores the role of online RAG, and Q5 addresses online performance.

\begin{table*}[h]
\centering
\small
\begin{tabular}{@{}lccccccccc@{}}
\toprule
\textbf{Model} & \multicolumn{5}{c}{\textbf{Domain Task (Visual-Language)}} & \multicolumn{2}{c}{\textbf{Language Only}} & \multicolumn{1}{c}{\textbf{Overall}} \\ 
\cmidrule(r){2-6} \cmidrule(r){7-8} \cmidrule(r){9-9}
   & \textbf{TS} & \textbf{CT} & \textbf{CR} & \textbf{VAE}  & \textbf{PDG} & \textbf{SA} & \textbf{TAE} & \textbf{Average}   \\ 
\midrule
Qwen-VL         &               0.442 &             0.758 &                 0.791 &             0.720&             0.610 &    \textbf{0.895}&    0.416 &            0.662   \\

+10\%~~~Data         &               0.532 &             0.769 &                 0.768 &             0.781&             0.629 &             0.871&     0.313 &            0.666   \\

+20\%~~~Data         &               0.596 &    \textbf{0.826} &                 0.733 &    \textbf{0.811}&             0.628 & {0.885}&     0.670 &            0.735   \\

+50\%~~ Data          & {0.610} &    {0.824}&     {0.799} & {0.809}&   \textbf{0.635} &             0.868&    {0.649} & {0.742}  \\

+100\%~Data    & \textbf{0.718} &                  0.822 &        \textbf{0.847} &             0.790&       {0.631} &             0.878&    \textbf{0.715} &    \textbf{0.771}    \\

\bottomrule
\end{tabular}
\caption{We compare the performance of domain-specific models trained with different proportions of the dataset (10\%, 20\%, 50\%, and 100\%) on various domain-specific tasks. These tasks include Topic Selection (TS), Content Tagging (CT), Category Recognition (CR), Vision-Based Product Attribute Extraction (VAE), Product Description Generation (PDG), Sentiment Analysis (SA) and Text-Based Product Attribute Extraction (TAE).}
\vspace{-10pt}
\end{table*}

\subsection{RQ1: Enhanced Domain-Specific Knowledge}
To evaluate the model's understanding of domain-specific knowledge, we compared its performance on C2C e-commerce tasks involving both language-only and visual-language hybrid modalities. As shown in Table 4, the domain-specific model significantly outperforms baseline across various metrics. Notably, the model shows substantial improvements in tasks such as e-commerce topic selection and category recognition, while the gains in sentiment analysis are relatively smaller. This can be attributed to the close alignment of sentiment classification with general tasks, as well as its superior baseline performance.

By truncating the training data to 10\%, 20\%, 50\%, and 100\% of the original dataset, we obtained different models. The model trained with the full dataset achieved the highest average accuracy, followed by the model trained with 50\% of the data. In the Topic Selection and Category Recognition tasks, The accuracy increased significantly with the amount of training data. For the Content Tagging and Vision-Based Product Attribute Extraction tasks, accuracy improved significantly after adding 20\% of the data, but showed minor fluctuations with further increases in training data beyond 20\%.

\subsection{RQ2: Enhanced Domain-Specific Style Generation Ability}
We also evaluated whether the model's generated listings exhibit domain-specific stylistic elements. Given that style preferences are subjective, human evaluation is the most reliable method. An experienced e-commerce annotator was tasked with comparing the linguistic style of listings generated by different models for the same product and casting votes. The results, presented in Table 5, indicate a significant preference for our model's outputs. In contrast, Qwen-VL's listings were often perceived as unnatural, verbose, and overly marketing-oriented, which is undesirable in C2C personal seller scenarios. We also experimented with various prompts for Qwen-VL to mitigate prompt-induced biases.

\subsection{RQ3: Retains General Capabilities}
We assessed the model's retention of general capabilities using well-established benchmarks such as MMBench \cite{liu2023mmbench} , MME\cite{fu2023mme}, and SeedBench\cite{li2024seed}, drawing reference from the work of LLaVA 1.5 and Qwen-VL. Our model outperforms LLaVA 1.5 and Qwen-VL on the MMBench task, and achieves performance closely comparable to LLaVA 1.5 on the MME task. However, it demonstrates relatively weaker performance on the SeedBench task.

On one hand, SeedBench focuses on detailed image analysis tasks, including scene understanding, instance identity, instance location, instance counting. In contrast, MMBench emphasizes overall image analysis, encompassing tasks such as image topic and attribute recognition. Our training samples are based on commonly used general-domain data and additionally incorporate e-commerce product understanding, encompassing tasks such as category recognition and product attribute extraction. From the perspective of the high-quality training samples, this demonstrates a greater improvement for MMBench compared to the SeedBench tasks.On the other hand, the difficulty of the tasks reveals that SeedBench is indeed more challenging, as detailed image analysis requires the model to possess strong pixel resolution, multi-object recognition capabilities, and spatial recognition skills. Our model still has space for improvement on these tasks. The generalization obtained from existing universal samples aids in enhancing both instruction-following abilities and image recognition capabilities. Therefore, we will continue to refine these abilities in our future work.

\begin{table}[H]
\centering

\begin{adjustbox}{width=0.78\columnwidth}
\begin{tabular}{@{}lcccccc@{}}
\toprule

\textbf{Model} & \textbf{Win:Loss} & \textbf{Win Rate}  \\
\midrule
Ours VS Qwen-VL & \textbf{948}:101 & 90.3\%   \\
\bottomrule
\end{tabular}
\end{adjustbox}

\caption{Model performance in description generation style on C2C domain based on human evaluation.}
\end{table}
\vspace{-20pt}

\begin{table}[H]
\centering

\begin{adjustbox}{width=0.78\columnwidth}
\begin{tabular}{@{}lcccccc@{}}
\toprule

\textbf{Model} & \textbf{MMBench(en/cn)} & \textbf{MME} & \textbf{SeedBench}\\
\midrule
LLaVA 1.5  & 65.2/57.3 & 1808.4 & \textbf{65.8} \\
Qwen-VL   & 61.8/56.3 & \textbf{1860.0} & 64.8 \\
Ours  & \textbf{71.5/65.5} & 1813.0 & 49.0 \\
\bottomrule
\end{tabular}
\end{adjustbox}
\caption{Performance of different models on open-source benchmarks to evaluate their general capabilities.}
\end{table}

\begin{table*}[h]
    \centering
    \small
    \begin{tabular}{>{\centering\arraybackslash}p{0.8cm} >{\centering\arraybackslash}p{1cm} >{\centering\arraybackslash}p{1cm} >{\centering\arraybackslash}p{1cm} >{\centering\arraybackslash}p{0.65cm} >{\centering\arraybackslash}p{0.65cm} >{\centering\arraybackslash}p{0.65cm} >{\centering\arraybackslash}p{0.65cm} >
    {\centering\arraybackslash}p{0.8cm} >{\centering\arraybackslash}p{0.9cm} >{\centering\arraybackslash}p{0.9cm} >{\centering\arraybackslash}p{0.9cm} >{\centering\arraybackslash}p{0.9cm} >{\centering\arraybackslash}p{0.9cm}}
        \toprule
        \multicolumn{3}{c}{\textbf{Unit}} & \multicolumn{1}{c}{\textbf{Human}} & \multicolumn{8}{c}{\textbf{Machine Auto Evaluation}} \\
        \cmidrule(lr){1-3} \cmidrule(lr){4-4} \cmidrule(lr){5-12}
        \textbf{Image} & \textbf{Category} & \textbf{Reference}  & \textbf{ACC} & \textbf{SIM} & \textbf{BLEU1} & \textbf{BLEU2} & \textbf{BLEU3} & \textbf{BLEU4} & \textbf{ROUGE1} & \textbf{ROUGE2}  & \textbf{ROUGEL} \\
        \midrule
        \checkmark &            &            & 0.36 & 0.633  & 0.132 & 0.027 & 0.009  & 0.003 & 0.155  & 0.034 & 0.153 \\
        \checkmark & \checkmark &            & 0.35 & 0.639  & 0.134 & 0.027 & 0.009  & 0.004 & 0.157  & 0.036 & 0.156 \\
        \checkmark &            & \checkmark & {0.74} & \textbf{0.720}  & {0.173} & \textbf{0.057} & \textbf{0.029}  & \textbf{0.018} & \textbf{0.216}  & \textbf{0.080} & {0.191} \\
        
        \checkmark & \checkmark & \checkmark & \textbf{0.75} & {0.718}  & \textbf{0.174} & {0.056} & {0.028}  & {0.016} & \textbf{0.216}  & {0.078} & \textbf{0.193} \\
        \bottomrule
    \end{tabular}
    \caption{Evaluation of component ablation effects in Retrieval-Augmented Generation Models}
    \vspace{-10pt}
\end{table*}

\subsection{RQ4:  RAG Can Alleviate Hallucinations}
We employed a combination of human and machine evaluations for this assessment.\\ 
\textbf{Key Attribute Evaluation:} Based on product photos, user-generated descriptions, and model-generated descriptions, evaluators are required to assess the accuracy of the attributes (e.g., brand, model) in the model outputs. Subsequently, we can compute the accuracy rate. \\
\textbf{Machine Automatic Evaluation:} The content generated by the model was compared to the user-written descriptions using metrics such as SIM, BLEU and ROUGE. 

\hspace{1em}The specific results are shown in Table 7. As opposed to only giving the image to the MLLMs, our model significantly improved all metrics.Especially in the human manual evaluation of attribute accuracy, there was a 105\% improvement. These enhancements can be attributed to RAG's ability to provide richer and more accurate reference information, which effectively mitigates hallucination. This indicates that the information obtained solely from product images is limited and necessitates supplementary references. On the other hand, the direct contribution of product categories is relatively minor. The primary function of category prediction is to obtain the relevant attributes template, thereby enhancing the controllability of the generation process in RAG.

\subsection{RQ5: Online A/B Test Results}
To evaluate the performance of the IPL system, we conducted online A/B testing. The objective was to measure the adoption rate of product descriptions generated by IPL and to compare the advantages over not using IPL.  Our experiments demonstrate a high user acceptance rate for our system: up to 72\% of users are willing to continue modifying the automatically generated descriptions to complete product listings, and over 32\% of users adopt more than 50\% of the generated content. Furthermore, products utilizing the auto-description generation feature exhibit a 5.6\% improvement in overall quality scores compared to similar products that do not use this feature. The product quality score, an internal metric used by the platform to assess product quality, is primarily calculated based on the richness of descriptions and the aesthetic authenticity of photos. The details of the quality score definition can be found in Appendix C.2.

\section{Related Work}

\textbf{Multimodal LLM:} Recent advances in large language models such as GPT-4, LLaMA\cite{touvron2023llama}, and Qwen\cite{bai2023qwen1}, have demonstrated impressive capabilities in understanding world knowledge and generating diverse text. These models have shown significant potential in zero-shot or few-shot\cite{wang2020generalizing} learning scenarios, exhibiting strong instruction-following abilities\cite{ouyang2022training}. Recent works, including BLIP-2\cite{li2023blip}, MiniGPT-4\cite{zhu2023minigpt}, and Qwen-VL\cite{bai2023qwen2}, have explored integrating visual and textual modalities from various perspectives. However, these models lack training on domain-specific (C2C) private data, resulting in insufficient domain understanding and inconsistent domain-specific style outputs, which limits their effectiveness in related tasks.\\
\textbf{Retrieval-Augmented Generation:}  Hallucination remains a major challenge in the development of LLMs\cite{guerreiro2023hallucinations}\cite{ji2023survey}. Approaches such as VisualGPT\cite{wu2023visual}, HuggingGPT\cite{shen2024hugginggpt}, and ToolFormer\cite{schick2024toolformer} leverage existing mature modules to perform complex operations. Another method, involves text retrieval-based augmentation\cite{guu2020retrieval,izacard2023atlas,robertson2009probabilistic,karpukhin2020dense}, where external resources\cite{guu2020retrieval} or web-retrieved\cite{nakano2021webgpt} texts are fed into the prompts to provide LLMs with more accurate \cite{mallen2022not,kandpal2023large}reference information to mitigate hallucinations\cite{li2022pre,kang2023impact}. Unlike these methods, our research uniquely integrates visual-based retrieval augmentation with MLLMs and successfully applies it in the e-commerce domain, addressing the hallucination problem while enhancing task-specific performance.

\section{Conclusion}

We presented IPL system, a novel framework that generates high-quality, accurate product descriptions based on images, enhancing item listing efficiency in the C2C market. By leveraging MLLMs trained via Domain Injection, our model gains deeper domain-specific knowledge and style compared to the original model (Qwen-VL). The implementation of Online RAG, which uses similar product images as reference, reduces hallucination in MLLMs, resulting in more precise descriptions. The effectiveness of our framework is demonstrated through human evaluations, machine assessments, and Online A/B testing.

\section{Limitations}

Our IPL system generates precise descriptions tailored to individual seller styles, streamlining the posting process and enhancing the quality of listings. Our system exhibits notable potential for further optimization. Firstly, the core attributes template is predominantly based on extensive descriptive statistics and do not yet account for personalized user posting styles. Secondly, the accuracy of generated descriptions for certain long-tail categories requires improvement. To advance our system, we intend to incorporate additional training samples from long-tail categories and integrate user personalization data. This approach aims to enhance the accuracy and personalization of product descriptions, thereby increasing adoption rates and aiding users in efficiently producing high-quality descriptions.

\bibliography{custom}

\begin{thebibliography}{40}
\providecommand{\natexlab}[1]{#1}

\bibitem[{Achiam et~al.(2023)Achiam, Adler, Agarwal, Ahmad, Akkaya, Aleman, Almeida, Altenschmidt, Altman, Anadkat et~al.}]{achiam2023gpt}
Josh Achiam, Steven Adler, Sandhini Agarwal, Lama Ahmad, Ilge Akkaya, Florencia~Leoni Aleman, Diogo Almeida, Janko Altenschmidt, Sam Altman, Shyamal Anadkat, et~al. 2023.
\newblock Gpt-4 technical report.
\newblock \emph{arXiv preprint arXiv:2303.08774}.

\bibitem[{Aminabadi et~al.(2022)Aminabadi, Rajbhandari, Awan, Li, Li, Zheng, Ruwase, Smith, Zhang, Rasley et~al.}]{aminabadi2022deepspeed}
Reza~Yazdani Aminabadi, Samyam Rajbhandari, Ammar~Ahmad Awan, Cheng Li, Du~Li, Elton Zheng, Olatunji Ruwase, Shaden Smith, Minjia Zhang, Jeff Rasley, et~al. 2022.
\newblock Deepspeed-inference: enabling efficient inference of transformer models at unprecedented scale.
\newblock In \emph{SC22: International Conference for High Performance Computing, Networking, Storage and Analysis}, pages 1--15. IEEE.

\bibitem[{Bai et~al.(2023{\natexlab{a}})Bai, Bai, Chu, Cui, Dang, Deng, Fan, Ge, Han, Huang et~al.}]{bai2023qwen1}
Jinze Bai, Shuai Bai, Yunfei Chu, Zeyu Cui, Kai Dang, Xiaodong Deng, Yang Fan, Wenbin Ge, Yu~Han, Fei Huang, et~al. 2023{\natexlab{a}}.
\newblock Qwen technical report.
\newblock \emph{arXiv preprint arXiv:2309.16609}.

\bibitem[{Bai et~al.(2023{\natexlab{b}})Bai, Bai, Yang, Wang, Tan, Wang, Lin, Zhou, and Zhou}]{bai2023qwen2}
Jinze Bai, Shuai Bai, Shusheng Yang, Shijie Wang, Sinan Tan, Peng Wang, Junyang Lin, Chang Zhou, and Jingren Zhou. 2023{\natexlab{b}}.
\newblock Qwen-vl: A frontier large vision-language model with versatile abilities.
\newblock \emph{arXiv preprint arXiv:2308.12966}.

\bibitem[{Chen et~al.(2024)Chen, Chen, Liu, Jiang, and Wang}]{chen2024humans}
Guiming~Hardy Chen, Shunian Chen, Ziche Liu, Feng Jiang, and Benyou Wang. 2024.
\newblock Humans or llms as the judge? a study on judgement biases.
\newblock \emph{arXiv preprint arXiv:2402.10669}.

\bibitem[{Dao(2023)}]{dao2023flashattention}
Tri Dao. 2023.
\newblock Flashattention-2: Faster attention with better parallelism and work partitioning.
\newblock \emph{arXiv preprint arXiv:2307.08691}.

\bibitem[{Dao et~al.(2022)Dao, Fu, Ermon, Rudra, and R{\'e}}]{dao2022flashattention}
Tri Dao, Dan Fu, Stefano Ermon, Atri Rudra, and Christopher R{\'e}. 2022.
\newblock Flashattention: Fast and memory-efficient exact attention with io-awareness.
\newblock \emph{Advances in Neural Information Processing Systems}, 35:16344--16359.

\bibitem[{Escursell et~al.(2021)Escursell, Llorach-Massana, and Roncero}]{escursell2021sustainability}
S{\'\i}lvia Escursell, Pere Llorach-Massana, and M~Blanca Roncero. 2021.
\newblock Sustainability in e-commerce packaging: A review.
\newblock \emph{Journal of cleaner production}, 280:124314.

\bibitem[{Fu et~al.(2023)Fu, Chen, Shen, Qin, Zhang, Lin, Yang, Zheng, Li, Sun, Wu, and Ji}]{fu2023mme}
Chaoyou Fu, Peixian Chen, Yunhang Shen, Yulei Qin, Mengdan Zhang, Xu~Lin, Jinrui Yang, Xiawu Zheng, Ke~Li, Xing Sun, Yunsheng Wu, and Rongrong Ji. 2023.
\newblock Mme: A comprehensive evaluation benchmark for multimodal large language models.
\newblock \emph{arXiv preprint arXiv:2306.13394}.

\bibitem[{Guerreiro et~al.(2023)Guerreiro, Alves, Waldendorf, Haddow, Birch, Colombo, and Martins}]{guerreiro2023hallucinations}
Nuno~M Guerreiro, Duarte~M Alves, Jonas Waldendorf, Barry Haddow, Alexandra Birch, Pierre Colombo, and Andr{\'e}~FT Martins. 2023.
\newblock Hallucinations in large multilingual translation models.
\newblock \emph{Transactions of the Association for Computational Linguistics}, 11:1500--1517.

\bibitem[{Guu et~al.(2020)Guu, Lee, Tung, Pasupat, and Chang}]{guu2020retrieval}
Kelvin Guu, Kenton Lee, Zora Tung, Panupong Pasupat, and Mingwei Chang. 2020.
\newblock Retrieval augmented language model pre-training.
\newblock In \emph{International conference on machine learning}, pages 3929--3938. PMLR.

\bibitem[{Herold et~al.(2024)Herold, Kozielski, Ekimov, Petrushkov, Vandenbussche, and Khadivi}]{herold2024lilium}
Christian Herold, Michael Kozielski, Leonid Ekimov, Pavel Petrushkov, Pierre-Yves Vandenbussche, and Shahram Khadivi. 2024.
\newblock Lilium: ebay's large language models for e-commerce.
\newblock \emph{arXiv preprint arXiv:2406.12023}.

\bibitem[{Izacard et~al.(2023)Izacard, Lewis, Lomeli, Hosseini, Petroni, Schick, Dwivedi-Yu, Joulin, Riedel, and Grave}]{izacard2023atlas}
Gautier Izacard, Patrick Lewis, Maria Lomeli, Lucas Hosseini, Fabio Petroni, Timo Schick, Jane Dwivedi-Yu, Armand Joulin, Sebastian Riedel, and Edouard Grave. 2023.
\newblock Atlas: Few-shot learning with retrieval augmented language models.
\newblock \emph{Journal of Machine Learning Research}, 24(251):1--43.

\bibitem[{Ji et~al.(2023)Ji, Lee, Frieske, Yu, Su, Xu, Ishii, Bang, Madotto, and Fung}]{ji2023survey}
Ziwei Ji, Nayeon Lee, Rita Frieske, Tiezheng Yu, Dan Su, Yan Xu, Etsuko Ishii, Ye~Jin Bang, Andrea Madotto, and Pascale Fung. 2023.
\newblock Survey of hallucination in natural language generation.
\newblock \emph{ACM Computing Surveys}, 55(12):1--38.

\bibitem[{Jiang et~al.(2024)Jiang, Jiang, Chu, Gulati, and Garg}]{jiang2024hallucination}
Ling Jiang, Keer Jiang, Xiaoyu Chu, Saaransh Gulati, and Pulkit Garg. 2024.
\newblock Hallucination detection in llm-enriched product listings.
\newblock In \emph{Proceedings of the Seventh Workshop on e-Commerce and NLP\@ LREC-COLING 2024}, pages 29--39.

\bibitem[{Kandpal et~al.(2023)Kandpal, Deng, Roberts, Wallace, and Raffel}]{kandpal2023large}
Nikhil Kandpal, Haikang Deng, Adam Roberts, Eric Wallace, and Colin Raffel. 2023.
\newblock Large language models struggle to learn long-tail knowledge.
\newblock In \emph{International Conference on Machine Learning}, pages 15696--15707. PMLR.

\bibitem[{Kang and Choi(2023)}]{kang2023impact}
Cheongwoong Kang and Jaesik Choi. 2023.
\newblock Impact of co-occurrence on factual knowledge of large language models.
\newblock \emph{arXiv preprint arXiv:2310.08256}.

\bibitem[{Karpukhin et~al.(2020)Karpukhin, O{\u{g}}uz, Min, Lewis, Wu, Edunov, Chen, and Yih}]{karpukhin2020dense}
Vladimir Karpukhin, Barlas O{\u{g}}uz, Sewon Min, Patrick Lewis, Ledell Wu, Sergey Edunov, Danqi Chen, and Wen-tau Yih. 2020.
\newblock Dense passage retrieval for open-domain question answering.
\newblock \emph{arXiv preprint arXiv:2004.04906}.

\bibitem[{Kwon et~al.(2023)Kwon, Li, Zhuang, Sheng, Zheng, Yu, Gonzalez, Zhang, and Stoica}]{kwon2023efficient}
Woosuk Kwon, Zhuohan Li, Siyuan Zhuang, Ying Sheng, Lianmin Zheng, Cody~Hao Yu, Joseph~E. Gonzalez, Hao Zhang, and Ion Stoica. 2023.
\newblock Efficient memory management for large language model serving with pagedattention.
\newblock In \emph{Proceedings of the ACM SIGOPS 29th Symposium on Operating Systems Principles}.

\bibitem[{Li et~al.(2024)Li, Ge, Ge, Wang, Wang, Zhang, and Shan}]{li2024seed}
Bohao Li, Yuying Ge, Yixiao Ge, Guangzhi Wang, Rui Wang, Ruimao Zhang, and Ying Shan. 2024.
\newblock Seed-bench: Benchmarking multimodal large language models.
\newblock In \emph{Proceedings of the IEEE/CVF Conference on Computer Vision and Pattern Recognition}, pages 13299--13308.

\bibitem[{Li et~al.(2023)Li, Li, Savarese, and Hoi}]{li2023blip}
Junnan Li, Dongxu Li, Silvio Savarese, and Steven Hoi. 2023.
\newblock Blip-2: Bootstrapping language-image pre-training with frozen image encoders and large language models.
\newblock In \emph{International conference on machine learning}, pages 19730--19742. PMLR.

\bibitem[{Li et~al.(2021)Li, Selvaraju, Gotmare, Joty, Xiong, and Hoi}]{li2021align}
Junnan Li, Ramprasaath Selvaraju, Akhilesh Gotmare, Shafiq Joty, Caiming Xiong, and Steven Chu~Hong Hoi. 2021.
\newblock Align before fuse: Vision and language representation learning with momentum distillation.
\newblock \emph{Advances in neural information processing systems}, 34:9694--9705.

\bibitem[{Li et~al.(2022)Li, Li, Shang, Dong, Sun, Liu, Ji, Jiang, and Liu}]{li2022pre}
Shaobo Li, Xiaoguang Li, Lifeng Shang, Zhenhua Dong, Chengjie Sun, Bingquan Liu, Zhenzhou Ji, Xin Jiang, and Qun Liu. 2022.
\newblock How pre-trained language models capture factual knowledge? a causal-inspired analysis.
\newblock \emph{arXiv preprint arXiv:2203.16747}.

\bibitem[{Liang et~al.(2022)Liang, Bommasani, Lee, Tsipras, Soylu, Yasunaga, Zhang, Narayanan, Wu, Kumar et~al.}]{liang2022holistic}
Percy Liang, Rishi Bommasani, Tony Lee, Dimitris Tsipras, Dilara Soylu, Michihiro Yasunaga, Yian Zhang, Deepak Narayanan, Yuhuai Wu, Ananya Kumar, et~al. 2022.
\newblock Holistic evaluation of language models.
\newblock \emph{arXiv preprint arXiv:2211.09110}.

\bibitem[{Lin(2004)}]{lin2004rouge}
Chin-Yew Lin. 2004.
\newblock Rouge: A package for automatic evaluation of summaries.
\newblock In \emph{Text summarization branches out}, pages 74--81.

\bibitem[{Liu et~al.(2023)Liu, Duan, Zhang, Li, Zhang, Zhao, Yuan, Wang, He, Liu et~al.}]{liu2023mmbench}
Yuan Liu, Haodong Duan, Yuanhan Zhang, Bo~Li, Songyang Zhang, Wangbo Zhao, Yike Yuan, Jiaqi Wang, Conghui He, Ziwei Liu, et~al. 2023.
\newblock Mmbench: Is your multi-modal model an all-around player?
\newblock \emph{arXiv preprint arXiv:2307.06281}.

\bibitem[{Mallen et~al.(2022)Mallen, Asai, Zhong, Das, Khashabi, and Hajishirzi}]{mallen2022not}
Alex Mallen, Akari Asai, Victor Zhong, Rajarshi Das, Daniel Khashabi, and Hannaneh Hajishirzi. 2022.
\newblock When not to trust language models: Investigating effectiveness of parametric and non-parametric memories.
\newblock \emph{arXiv preprint arXiv:2212.10511}.

\bibitem[{Nakano et~al.(2021)Nakano, Hilton, Balaji, Wu, Ouyang, Kim, Hesse, Jain, Kosaraju, Saunders et~al.}]{nakano2021webgpt}
Reiichiro Nakano, Jacob Hilton, Suchir Balaji, Jeff Wu, Long Ouyang, Christina Kim, Christopher Hesse, Shantanu Jain, Vineet Kosaraju, William Saunders, et~al. 2021.
\newblock \href {https://arxiv.org/abs/2112.09332} {Webgpt: Browser-assisted question-answering with human feedback, 2021}.
\newblock \emph{arXiv preprint arXiv:2112.09332}.

\bibitem[{Ouyang et~al.(2022)Ouyang, Wu, Jiang, Almeida, Wainwright, Mishkin, Zhang, Agarwal, Slama, Ray et~al.}]{ouyang2022training}
Long Ouyang, Jeffrey Wu, Xu~Jiang, Diogo Almeida, Carroll Wainwright, Pamela Mishkin, Chong Zhang, Sandhini Agarwal, Katarina Slama, Alex Ray, et~al. 2022.
\newblock Training language models to follow instructions with human feedback.
\newblock \emph{Advances in neural information processing systems}, 35:27730--27744.

\bibitem[{Papineni et~al.(2002)Papineni, Roukos, Ward, and Zhu}]{papineni2002bleu}
Kishore Papineni, Salim Roukos, Todd Ward, and Wei-Jing Zhu. 2002.
\newblock Bleu: a method for automatic evaluation of machine translation.
\newblock In \emph{Proceedings of the 40th annual meeting of the Association for Computational Linguistics}, pages 311--318.

\bibitem[{Perez and Ribeiro(2022)}]{perez2022ignore}
F{\'a}bio Perez and Ian Ribeiro. 2022.
\newblock Ignore previous prompt: Attack techniques for language models.
\newblock \emph{arXiv preprint arXiv:2211.09527}.

\bibitem[{Poerner et~al.(2019)Poerner, Waltinger, and Sch{\"u}tze}]{poerner2019bert}
Nina Poerner, Ulli Waltinger, and Hinrich Sch{\"u}tze. 2019.
\newblock E-bert: Efficient-yet-effective entity embeddings for bert.
\newblock \emph{arXiv preprint arXiv:1911.03681}.

\bibitem[{Robertson et~al.(2009)Robertson, Zaragoza et~al.}]{robertson2009probabilistic}
Stephen Robertson, Hugo Zaragoza, et~al. 2009.
\newblock The probabilistic relevance framework: Bm25 and beyond.
\newblock \emph{Foundations and Trends{\textregistered} in Information Retrieval}, 3(4):333--389.

\bibitem[{Schick et~al.(2024)Schick, Dwivedi-Yu, Dess{\`\i}, Raileanu, Lomeli, Hambro, Zettlemoyer, Cancedda, and Scialom}]{schick2024toolformer}
Timo Schick, Jane Dwivedi-Yu, Roberto Dess{\`\i}, Roberta Raileanu, Maria Lomeli, Eric Hambro, Luke Zettlemoyer, Nicola Cancedda, and Thomas Scialom. 2024.
\newblock Toolformer: Language models can teach themselves to use tools.
\newblock \emph{Advances in Neural Information Processing Systems}, 36.

\bibitem[{Shen et~al.(2024)Shen, Song, Tan, Li, Lu, and Zhuang}]{shen2024hugginggpt}
Yongliang Shen, Kaitao Song, Xu~Tan, Dongsheng Li, Weiming Lu, and Yueting Zhuang. 2024.
\newblock Hugginggpt: Solving ai tasks with chatgpt and its friends in hugging face.
\newblock \emph{Advances in Neural Information Processing Systems}, 36.

\bibitem[{Touvron et~al.(2023)Touvron, Lavril, Izacard, Martinet, Lachaux, Lacroix, Rozi{\`e}re, Goyal, Hambro, Azhar et~al.}]{touvron2023llama}
Hugo Touvron, Thibaut Lavril, Gautier Izacard, Xavier Martinet, Marie-Anne Lachaux, Timoth{\'e}e Lacroix, Baptiste Rozi{\`e}re, Naman Goyal, Eric Hambro, Faisal Azhar, et~al. 2023.
\newblock Llama: Open and efficient foundation language models.
\newblock \emph{arXiv preprint arXiv:2302.13971}.

\bibitem[{Wang et~al.(2020)Wang, Yao, Kwok, and Ni}]{wang2020generalizing}
Yaqing Wang, Quanming Yao, James~T Kwok, and Lionel~M Ni. 2020.
\newblock Generalizing from a few examples: A survey on few-shot learning.
\newblock \emph{ACM computing surveys (csur)}, 53(3):1--34.

\bibitem[{Wu et~al.(2023)Wu, Yin, Qi, Wang, Tang, and Duan}]{wu2023visual}
Chenfei Wu, Shengming Yin, Weizhen Qi, Xiaodong Wang, Zecheng Tang, and Nan Duan. 2023.
\newblock Visual chatgpt: Talking, drawing and editing with visual foundation models.
\newblock \emph{arXiv preprint arXiv:2303.04671}.

\bibitem[{Zhang et~al.(2018)Zhang, Pan, Zheng, Zhao, Zhang, Ren, and Jin}]{zhang2018visual}
Yanhao Zhang, Pan Pan, Yun Zheng, Kang Zhao, Yingya Zhang, Xiaofeng Ren, and Rong Jin. 2018.
\newblock Visual search at alibaba.
\newblock In \emph{Proceedings of the 24th ACM SIGKDD international conference on knowledge discovery \& data mining}, pages 993--1001.

\bibitem[{Zhu et~al.(2023)Zhu, Chen, Shen, Li, and Elhoseiny}]{zhu2023minigpt}
Deyao Zhu, Jun Chen, Xiaoqian Shen, Xiang Li, and Mohamed Elhoseiny. 2023.
\newblock Minigpt-4: Enhancing vision-language understanding with advanced large language models.
\newblock \emph{arXiv preprint arXiv:2304.10592}.

\end{thebibliography}

\appendix

\clearpage 
\section{Data Processing}
\label{sec:appendix1}

Our model's training data comprises tasks related to product description generation, e-commerce domain understanding, and general capability tasks. The methods for collecting and constructing data for each type of task vary accordingly.

\subsection{Description Generation Data}
\label{sec:appendix2}

Product description generation is the core task of our model, with the goal of generating product descriptions in the style of C2C platforms based on user-uploaded images. To achieve this goal, the best data source can be considered as the products posted by actual users on the platform. Given the varying quality of user-posted products, data selection and cleaning are also crucial. Additionally, it is necessary to construct various description generation instructions to increase the richness and controllability of product description generation.
Data cleaning and selection include the following key steps:
\begin{itemize}
    \item First, filter out low-quality products based on product quality scores, which mainly consider the completeness of basic descriptions and the aesthetic quality of product photos;
    \item Filter out products with negative risks present on the platform, such as low-priced traffic attraction, traffic attraction to other platforms, and potential fraudulent products;
    \item Use a self-developed image-text matching model similar to CLIP to filter out products with low similarity between photos and descriptions;
    \item Apply heuristic rules to exclude products that do not meet generation standards, such as excessively long or short descriptions, inclusion of user privacy information, or special characters;
    \item Finally, perform stratified sampling based on categories to obtain training sample candidates with balanced categories.
\end{itemize}
For the diversity of instructions, we mainly provide three types of instructions: generating product descriptions directly from images, generating descriptions based on images + core attributes template, and generating product descriptions based on product images + core attributes template + reference information. Examples of the three types of instructions and model responses are shown in Table 8.

For generating product descriptions directly from images, we can directly format the cleaned product image and description pairs as instructions. For the second type of task, we need to first perform core attribute extraction on the target product descriptions and then concatenate the extracted attribute names as part of the description generation instructions to obtain the corresponding format of training data. Similarly, based on the second type of instructions, we include the extracted attribute values as part of the reference information within the instruction prompt, thus obtaining the third type of instruction tuning data.

\begin{table*}[t!]
  \begin{minipage}{0.8\linewidth}
\centering
\scalebox{0.88}{

\begin{tabular}{p{3.7cm} p{13.3cm} }
\toprule
 \multicolumn{2}{l}{\bf Instruction Design for Product Description Generation:}  \\
\midrule
&  \includegraphics[height=3cm]{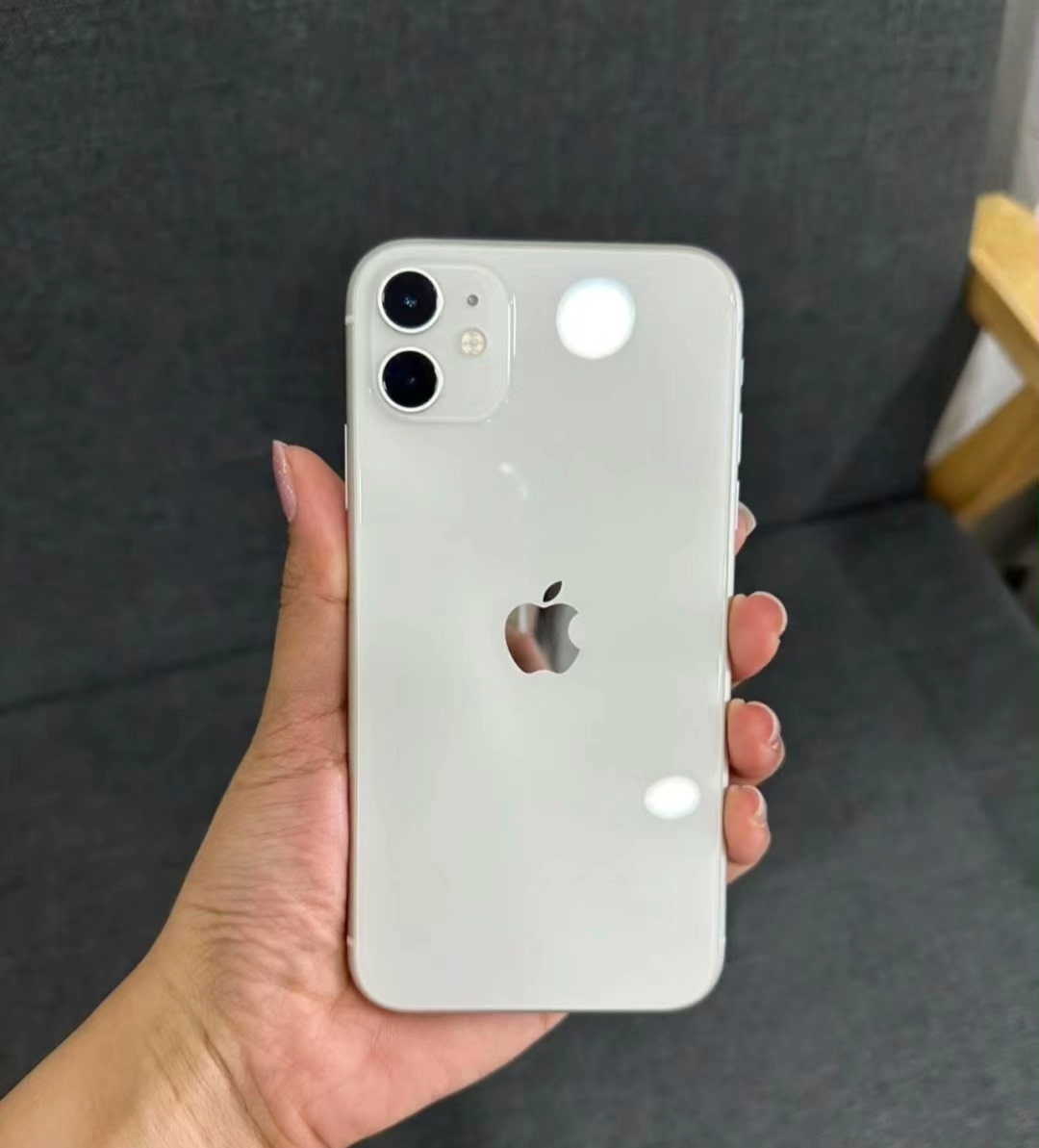} \\

User Description \\(Generation Target) & Apple iPhone 11, China version, 256GB, Silver, 90\% new, purchased from the official website. If interested, please contact me privately. \\
\midrule
Attribute Extraction \\ Results & 
\{
"Brand": "Apple", "Model": "iPhone 11", "Version Type": "China Version", "Memory Capacity": "256GB", "Color": "Silver", "Condition": "90\% New", "Purchase Channel": "Official Website"
\} 
\\
\midrule
Instruction Instance \\(generating product descriptions directly from images) & You are an experienced seller on a second-hand trading platform and need to post a listing for a mobile phone product. The product images are as shown. Please write a product description for this item, enhancing and expanding it reasonably.\\

\midrule
Instruction Instance \\(generating descriptions based on images + core attribute templates) & You are an experienced seller on a second-hand trading platform and need to post a listing for a mobile phone product. The product images are as shown.The copywriting template is Brand+Model+Version Type+Memory Capacity+Color+Condition+Purchase Channel. 
Please write a product description for this item, enhancing and expanding it reasonably according to the template.\\

\midrule
Instruction Instance \\(generating product descriptions based on product images + core attribute templates + reference information.) & You are an experienced seller on a second-hand trading platform and need to post a listing for a mobile phone product. The product images are as shown.The copywriting template is Brand+Model+Version Type+Memory Capacity+Color+Condition+Purchase Channel, where Brand is Apple, Model is iPhone 11, and Memory Capacity is 256GB.
Please write a product description for this item, enhancing and expanding it reasonably according to the template.\\

\bottomrule
\end{tabular}
}
\captionof{table}{Examples of different instruction designs for product description generation.}
\label{tab:visual_writing_task}
  \end{minipage}
\end{table*}

\subsection{E-commerce Understanding Data}
\label{sec:appendix3}

Introducing e-commerce domain task data aims to enhance the model's understanding of e-commerce knowledge, particularly the unique data distribution of C2C e-commerce platforms. To ensure the diversity of this data, we collect metadata based on two dimensions: technical direction and specific task type. The technical directions include classic product understanding on e-commerce platforms, search query understanding, relevance matching, data mining, and e-commerce QA, etc., while the task types include classification tasks, matching tasks, ranking tasks, and sequence labeling tasks. 

Additionally, our domain task data are all derived from the platform’s historically accumulated data, all of which have been manually annotated or ensured by other accuracy assurance methods to guarantee data quality. Finally, all the metadata are converted into instruction format for model training.

\subsection{General Instruction Data}
\label{sec:appendix4}
Training a model solely on domain-specific tasks induces overfitting to the instructions within the training data, thereby diminishing the model's generalization capability and its ability to follow general instructions. To mitigate this issue, we incorporated general task data into the training dataset, primarily sampling from the open-source data provided by the LLaVA1.5 project.

Since high-quality open-source data are typically in English, to enhance the model's performance in the native language and adapt to the platform's own data distribution, we automatically generate general instruction QA data using large language models for product photos. Specifically, for each product photo, we utilize a large language model to generate multiple potential instruction questions and their corresponding answers. Table 9 provides an example of the prompt engineering process utilized in this step. To further improve the accuracy of the generated answers, for each instruction question, we use a robust large language model to generate answers based on the given picture and instruction, thereby producing the final training data.

\begin{table*}[t!]
  \begin{minipage}{0.99\linewidth}
\centering
\scalebox{0.88}{
\begin{tabular}{p{3.7cm} p{13.3cm} }
\toprule
 \multicolumn{2}{l}{\bf Prompt Engineering for Automatic Generation of General QA Data:}  \\
\midrule
&  \includegraphics[height=3cm]{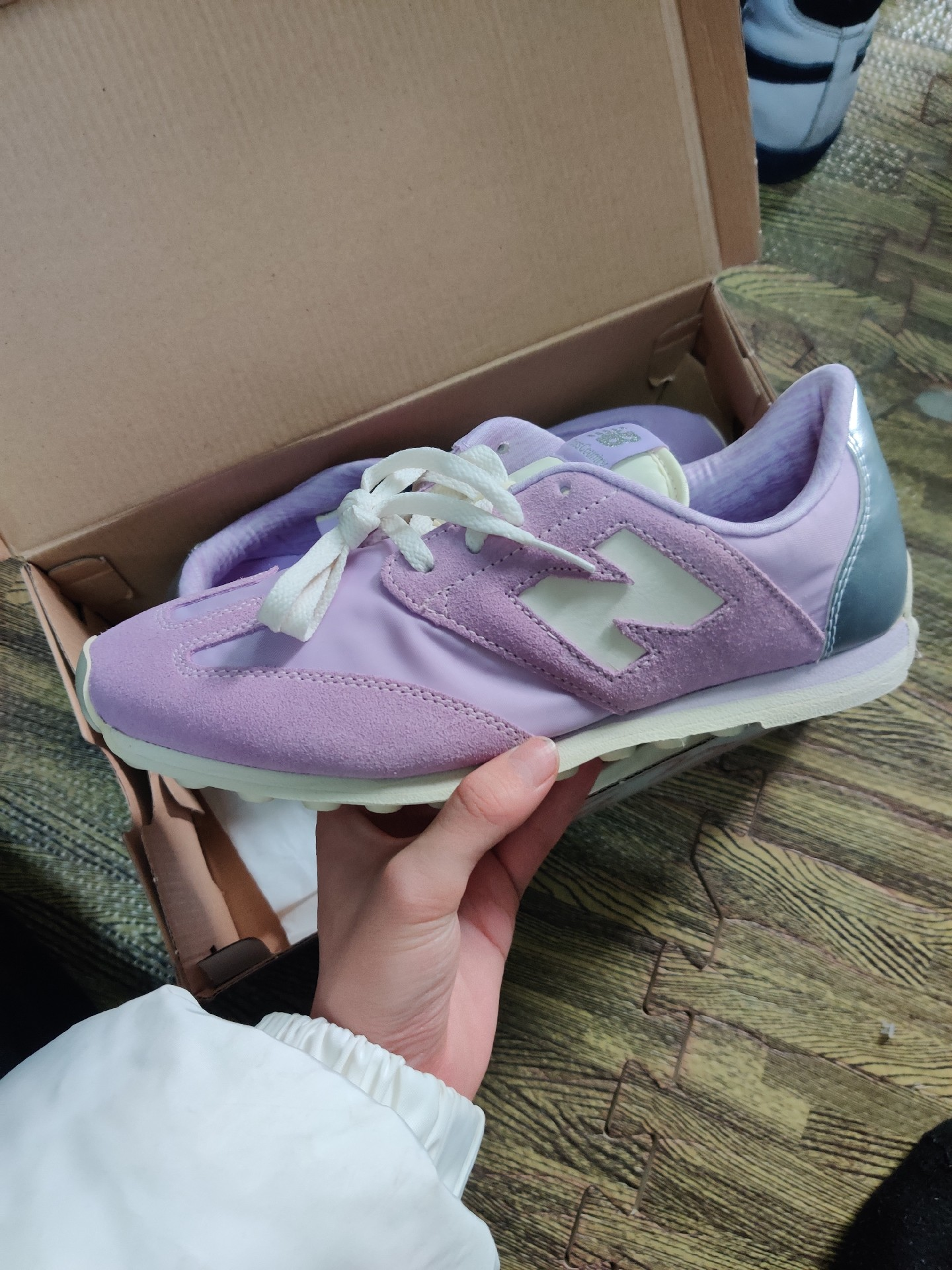} \\

Product Photo & \newshortname{} \\
\midrule
Prompt & 
Based on the given image, design multiple types of task questions and answers. The task types include: image information description, image emotion analysis, image action recognition, existence check of elements in the image, image text extraction, analysis of object interactions in the image, object attribute recognition in the image, image multiple-choice question answering, visual reasoning, visual common sense reasoning, image style appreciation, content creation based on the image, and writing product descriptions based on the image. Below, provide up to 20 diverse instructions for all the above tasks, including different language styles and precise answers. The instructions should include both questions and statements. Answers should be less than 100 words. Each task should have fewer than 3 instructions. Output format:

Instruction1: Example Instruction1

Answer1: Example Answer1

Task1: Example Task1

Instruction2: Example Instruction2

Answer2: Example Answer2

Task2: Example Task2

...\\
\midrule
Response & 
Instruction1: Please describe the objects in the image.  

Answer1: The image features a pair of purple sneakers placed in a cardboard box, with white shoelaces.  

Task1: Image Information Description  

Instruction2: How does the color of these shoes make you feel?  

Answer2: The purple color of the shoes gives a gentle and refreshing feeling.  

Task2: Image Emotion Analysis  

Instruction3: What is the person in the image doing?  

Answer3: The person in the image is holding the shoes with their hands, as if they are displaying or inspecting the shoes.  

Task3: Image Action Recognition  

Instruction4: Is there a label shown in the image?  

Answer4: No, there is no label visible in the image.  

Task4: Existence Check of Elements in the Image

...

Instruction9: Based on the image, what activities are these shoes suitable for?  

Answer9: Since they are sneakers, they are suitable for sports or casual activities.  

Task9: Visual Reasoning  

Instruction10: What are such sneakers usually used for?  

Answer10: Such sneakers are typically used for running, workouts, or casual daily wear.  

Task10: Visual Common Sense Reasoning  

Instruction11: What do you think of the design style of these shoes?  

Answer11: The design style of these shoes is modern and minimalist, with gentle colors that make them look comfortable and pleasant.  

Task11: Image Style Appreciation  

Instruction12: Based on the image, write a short story about these shoes.  

Answer12: One day, Xiaoming found his dream purple sneakers. He wore them and won the school marathon championship.  

Task12: Content Creation Based on the Image\\
\bottomrule
\end{tabular}
}
\captionof{table}{Sample of automatically generated general instructional QA data based on product photos using Large Language Model prompt engineering.}
\label{tab:visual_writing_task}
  \end{minipage}
\end{table*}

\section{Details of Training}
\label{sec:appendix5}

\subsection{Data Format of Supervised Fine-tuning}
\label{sec:appendix6}

Regarding the format of the training data, we follow the approach of Qwen-VL, converting the prepared instruction tuning data into ChatML (OpenAI) format, marking each interaction statement with special tokens (\textless im\_start\textgreater{} and \textless im\_end\textgreater{}) to denote dialogue termination. Training objectives focused on classic next token generation for language model optimization, excluding prompt prefixes and emphasizing special markers and model outputs (depicted in Table 10).

\begin{table}[H]
\centering
{\footnotesize 
\begin{tabular}{|>{\color{codeblack}\ttfamily}m{7cm}|} 
\hline
\textcolor{codeblue}{\textbf{The Dataset Format Example of ChatML}} \\ \hline
\textcolor{codeblue}{<im\_start>user} \\
\textcolor{codeblack}{Picture 1: <img>vg/VG\_100K\_2/649.jpg</img>What is the sign in the picture?<im\_end>} \\
\textcolor{codeblack}{<im\_start>assistant} \\
\textcolor{codeblue}{The sign is a road closure with an orange rhombus.<im\_end>} \\
\textcolor{codeblack}{<im\_start>user} \\
\textcolor{codeblack}{How is the weather in the picture?<im\_end>} \\
\textcolor{codeblack}{<im\_start>assistant} \\
\textcolor{codeblue}{The shape of the road closure sign is an orange rhombus.<im\_end>} \\ \hline
\end{tabular}
} 
\caption{Instruction Fine-Tuning data format.}
\end{table}

\subsection{Training Hyperparameters}
\label{sec:appendix7}

Table 11 presents some of the parameter settings used in the training process of our domain-specific model.

\begin{table}[H]
\centering

\begin{tabular}{@{}lc@{}}
\toprule
\textbf{Parameters} & \textbf{Value} \\
\midrule
ViT init & Qwen-VL-Chat \\
LLM init & Qwen-VL-Chat \\
VL Adapter init & Qwen-VL-Chat \\
Image resolution & 448x448 \\
ViT sequence length & 1024 \\
LLM sequence length & 1024 \\
Learnable query number & 256 \\
Learning rate & 1e-5 \\
Epoch & 3 \\
Training steps & 4788 \\
Learning rate schedule & Cosine decay \\
Global batch size & 768 \\
Gradient accumulation & 16 \\
Numerical precision & BF16 \\
DeepSpeed & Stage1 \\
\bottomrule
\end{tabular}

\caption{Parameter settings used in the training process.}

\end{table}

We employ the DeepSpeed ZeRO stage 1 approach for parallel training, utilizing 24 A800 GPUs to train on 1.27M data for 3 epochs, taking 16 hours, with an average throughput of 2.5 samples per second per GPU. We use the AdamW optimizer with $\beta_1 = 0.9$, $\beta_2 = 0.98$, and $\epsilon = 1 \times 10^{-6}$. We also apply a cosine learning rate schedule with a warmup ratio of 0.01.

\section{Details of Internal Evaluation Method}
\label{sec:appendix8}

\subsection{Evaluation Metrics for Visual Retrieval}
\label{sec:appendix9}

We evaluate the effectiveness of visual retrieval by assessing whether the query image and the retrieved images are identical or similar. Specifically, identical products refers to that the two items share the same SKU (Stock Keeping Unit), where both the key attributes (such as product name, brand, mode, etc.) and non-key attributes (such as color, size, etc.) must be exactly the same. Similar products stands for that the two items with the same SPU (Standard Product Unit), where only the key attributes is asked to be matched, leading to a significantly high accuracy compared to the same level. 

In our experiments, we found that similarity at the SPU level can provide accurate essential attributes, which significantly aids in the generation of final description.

\subsection{Calculation of product quality score}
\label{sec:appendix10}

Product quality score is computed using an explainable-and-linearly weighted formula based on the content description. Key features include categories, attributes, descriptions, images, videos and price. The weight for each feature is determined by professional operators based on the importance of each of the above-mentioned dimension. The formula is listed in the below.

\begin{equation}
    quality\_score = \sum_{i=1}^{N} w_i * feature_i
\end{equation}

where \( feature \) denotes the characteristics considered for quality score, such as the accuracy of the category, the attribute filling rate and the fluency of the description. \( w \) represents the weight assigned to each corresponding feature, and \( N \) is the total number of features, which in this case is 11.

\section{Error Detection and Exception Handling in Online Services}
\label{sec:appendix11}

We designed a set of exception-handling mechanisms over multiple stages for better accommodating the production system. 

During the input stage, the uploaded images may contain non-compliant content, such as prohibited products, pornography, and etc. To avoid such cases, we applied several machine learning models for security check, which can provide proper guideline when such harmful content has been identified.

In the pipeline stage, exceptions may also occur from different sub-modules, such as empty category prediction, empty visual search results, and etc. All of them would change the reference information of the MLLM input. To address such issue, we designed instructions that cover all of those cases during model training (more details in Table 8). In the worst case, the model is allowed to generate product descriptions solely based on the uploaded image. For instance, if the image search yields no results, the MLLM will utilize the image information, along with domain knowledge, to generate product description. It is worth noting that, the chance of hallucination increases in this case (refer to Table 7).

During the output stage, in the process of streaming output, we keep monitoring content safety. Once the harmful content is detected, the content generation process will be halted, with an subsequent notification to the user for modification. Additionally, if the output exceeds the pre-defined content length limit, we will automatically truncate it to avoid system failure.

Lastly, in the case of request timeout, we keep the existing product listing function intact, allowing users to manually edit the content description.

\section{Summary of the In-house Evaluation Benchmarks}
\label{sec:appendix12}

In the experimental section, we designed multiple in-house validation datasets to evaluate the domain adaptation capabilities of our model. All data were sourced from real e-commerce scenarios and the target labels were either manually annotated or confirmed by actual platform users, then converted into instruction format. Table 12 presents the various evaluation datasets along with their evaluation details.

A unified test instruction is used for the evaluation tasks without special optimizations for the model. Additionally, some tasks will provide a few-shot examples to ensure the model outputs answers in the expected format. For the calculation of evaluation metrics, we use string matching to determine whether the generated results are consistent with the target answers. Manual verification has shown that this method has extremely high accuracy in our evaluation task.

\begin{table*}[ht]
    \centering
    \small 
    \begin{adjustbox}{width=2.0\columnwidth}
    \begin{tabular}{p{1.9cm}|p{1.6cm}|p{5cm}|p{0.5cm}|p{1.2cm}}
        \toprule
         Task & Dataset & Description & Size & Metric  \\
         \midrule
         Multi-choice Question &  Topic Selection (TS) & Given the content of a post published by a user in e-commerce scenario, along with a set of candidate topics, the objective is to select the topic from the candidate set that matches the post content. & 5k & Accuracy($\uparrow$) \\
         \midrule
         Multi-choice Question & Content Tagging (CT) & Given the content of a user's post in an e-commerce scenario and a set of candidate categorys, the objective is to select the category from the candidate set that match the post content. & 5k & Accuracy($\uparrow$) \\
         \midrule
         Multi-choice Question & Category Recognition (CR) & Given the product image and text information posted by users in the e-commerce scenario, as well as the candidate category set, the goal is to select the category to which the product belongs. & 5k & Accuracy($\uparrow$) \\
         \midrule
         Multi-choice Question & Vision-Based Product Attribute Extraction (VAE) & Given a product photo, the desired attribute, and a list of candidate attribute values, the goal is to select the correct attribute value from the candidate list. & 5k & Accuracy($\uparrow$) \\
         \midrule
         Image Caption & Product Description Generation (PDG) &Given user-uploaded product photos, the goal is to generate corresponding product descriptions that closely match the content written by the users themselves. & 2k & SIM($\uparrow$) \\
         \midrule
         Multi-choice Question & Sentiment Analysis (SA) & After purchasing products, users provide feedback on their buying experience. The objective is to distinguish whether the user's review is positive or negative. & 5k & Accuracy($\uparrow$) \\
         \midrule
         Information Extraction & Text-Based Product Attribute Extraction (TAE) & Given the textual description of a product and the list of desired attributes to be extracted, the objective is to extract the corresponding attribute values from the description text. & 5k & Accuracy($\uparrow$) \\

         \bottomrule
    \end{tabular}
    \end{adjustbox}
    \caption{Summary of the domain evaluation benchmarks.}
    \label{tab:benchmark}
\end{table*}

\end{document}